\documentclass[letterpaper, 10pt, journal, twoside]{IEEEtran}

\IEEEoverridecommandlockouts
\usepackage{cite}
\usepackage{amsmath,amssymb,amsfonts}

\usepackage{algorithmic}
\usepackage{graphicx}
\usepackage{subcaption}
\usepackage{textcomp}
\usepackage[table]{xcolor}
\usepackage{booktabs}
\usepackage{multirow}
\usepackage[colorlinks=false]{hyperref}
\usepackage[normalem]{ulem}

\def\BibTeX{{\rm B\kern-.05em{\sc i\kern-.025em b}\kern-.08em
    T\kern-.1667em\lower.7ex\hbox{E}\kern-.125emX}}

\begin{document}

\newcommand{\tpm}{\scriptstyle\pm}

\newcommand{\minew}[1]{{\color{blue}{#1}}}
\newcommand{\miold}[1]{{\color{red}\sout{#1}}}
% the below two commands are defined to cancel the highlight and compile the clean version
%\newcommand{\minew}[1]{{\color{black}{#1}}}
%\newcommand{\miold}[1]{\iffalse{#1}\fi}

\title{
MAPF-World: Action World Model for Multi-Agent Path Finding
}

\author{
Zhanjiang Yang$^{1,2*}$,
Yueming Li$^{1*}$,
Yang Shen$^{3}$,
Meng Li$^1$, and
Lijun Sun$^{\dagger1}$,~\IEEEmembership{Member,~IEEE}
% \thanks{$^{*}$ Equal contribution.}
\thanks{Manuscript received: February 6, 2026; Revised:
May 31, 2026; Accepted: July 27, 2026.}
\thanks{This paper was recommended for publication by Editor M. Ani Hsieh upon evaluation of the Associate Editor and Reviewers' comments.}
\thanks{This work was supported by Natural Science Foundation of Top Talent of SZTU under Grant No. GDRC202521.}
\thanks{$^1$Shenzhen Technology University, China.}
\thanks{$^2$Shenzhen University, China.}
\thanks{$^3$University of Technology Sydney, Australia.}
\thanks{$^*$Equal contribution.}
\thanks{$^{\dagger}$Corresponding author: {\tt\small sunlijun@sztu.edu.cn}.}
\thanks{Digital Object Identifier (DOI): see top of this page.}
}

\markboth{IEEE Robotics and Automation Letters. Preprint Version. Accepted August, 2026}%
{Yang \MakeLowercase{\textit{et al.}}: MAPF-World: Action World Model for Multi-Agent Path Finding}
\maketitle

\begin{abstract}
Multi-agent path finding (MAPF) studies the problem of planning conflict-free paths for multiple agents from given start locations to designated goals, with applications in robot-assisted logistics and social navigation. Recent decentralized learned solvers have shown promise for large-scale MAPF, particularly when leveraging foundation models and large datasets. However, most existing methods rely on reactive policies, often resulting in congestion, deadlocks, and degraded generalization in high agent-density environments. To address these limitations, we propose MAPF-World, an autoregressive action world model for MAPF that unifies short-horizon local future prediction and action generation, enabling decision-making beyond immediate local observations. 
MAPF-World models short-horizon local dynamics by predicting the next local observation and neighboring agents' action intentions, capturing both spatial structures and temporal interaction patterns.
We further introduce a spatio-agent positional encoding that integrates spatial awareness with agent-level semantics in Transformer-based architectures, facilitating more coordinated multi-agent behaviors. In addition, we augment existing MAPF benchmarks by introducing an automated map generator grounded in real-world urban layouts, aiming to narrow the gap between synthetic simulation and practical deployment scenarios. Extensive experiments across diverse map types and interaction settings demonstrate that MAPF-World achieves strong performance compared with existing learned solvers. Notably, it exhibits robust zero-shot generalization and maintains a high success rate even as agent density increases. 
\end{abstract}



\section{Introduction}
Multi-agent path finding (MAPF)~\cite{stern2019mapf} is a fundamental problem in robotics and embodied intelligence, with diverse applications in autonomous driving, collective robotics, indoor service robots, factory robots, and warehouse automation~\cite{sun2025multi}. With the increasing deployment of multi-agent systems in these real-world scenarios, MAPF has attracted growing research attention~\cite{li2025proactive,wang2025multi}. The problem is known to be NP-hard \cite{yu2013structure,yu2015intractability,ma2016multi,banfi2017intractability}, presenting significant challenges due to strict conflict-free constraints on path planning, especially in large-scale scenarios with many agents and complex layouts under real-time requirements.
Moreover, while existing MAPF benchmarks include realistic maps, these layouts are typically fixed and lack an automated generation process, which constrains the scale and diversity of evaluation and hampers robust generalization comparisons.

\begin{figure}[htbp]
    \centering
    \includegraphics[width=0.92\linewidth]{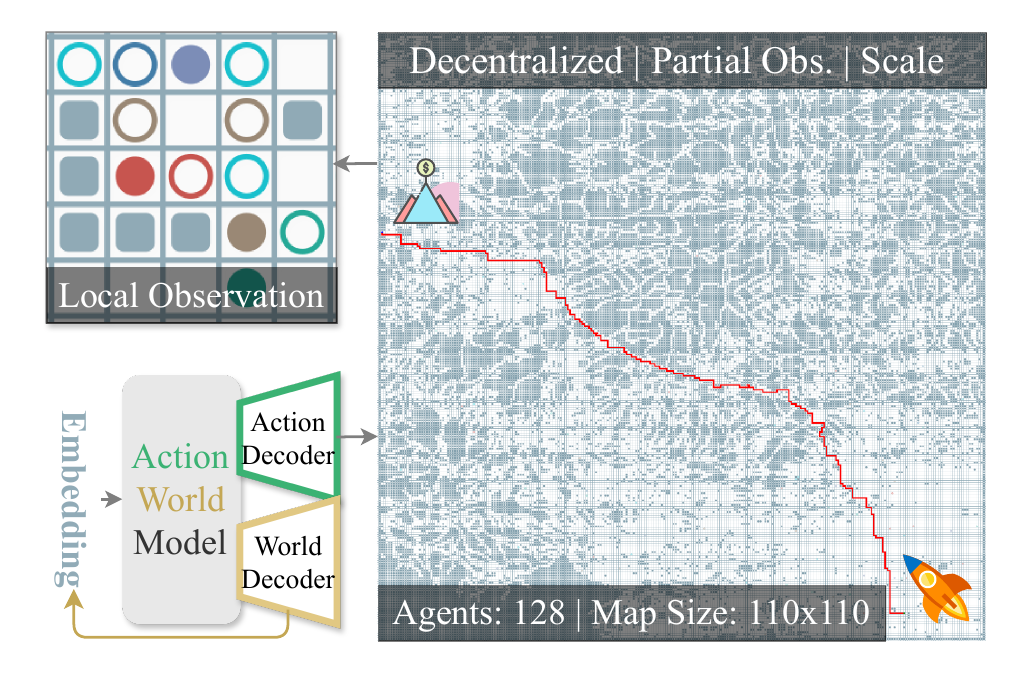}
    \caption{\textbf{MAPF-World} is a decentralized action--world model for multi-agent path finding (MAPF) under partial observation. Its action branch generates actions, while its world branch predicts short-horizon future local observations to support coordination in crowded and structurally complex scenarios.}
    \label{fig:teaser}
\end{figure}

From an algorithmic perspective, classical centralized and heuristic-based methods, such as CBS~\cite{sharon2015conflict} and PBS~\cite{liu2019task}, provide (sub-)optimal solutions but struggle with scalability. To address this, PIBT~\cite{okumura2022priority} offers a heuristic online approach with compromised solution quality, while LaCAM*~\cite{okumura2023improving,okumura2023engineering} presents an anytime near-optimal method that excels in efficiency. As emerging solutions, decentralized learning-based methods primarily leverage imitation learning (IL) and reinforcement learning (RL), enhancing performance by incorporating additional information through communication, centralized collision resolution, and global guidance, as seen in SCRIMP~\cite{wang2023scrimp} and SILLM~\cite{jiang2024deploying}. More recent learnable solvers push the frontier further by utilizing large datasets, foundation models, and fine-tuning techniques, such as MAPF-GPT~\cite{andreychuk2025mapf} and MAPF-GPT-DDG~\cite{andreychuk2025advancing}. 
% Additionally, W. Anzhe et al.~\cite{wang2025multi} explore topological formulations with graph-aware reinforcement learning to alleviate computational costs and improve scalability. 
However, these methods design reactive policy models that produce actions solely based on current observations, potentially failing to capture the complexity of general environmental layouts and temporal dynamics of crowded agents.
As a result, performance can degrade due to congestion, deadlocks, and livelocks, particularly in complex, high-density, or long-horizon planning scenarios under partial observation.

To overcome the limitations of reactive policies, we propose {\bf MAPF-World}\footnote{https://github.com/Fibonaccirabbit/MPAF-World}, an autoregressive \textbf{action world model} for MAPF inspired by Kahneman’s fast--slow dual-system theory~\cite{kahneman2011thinking}. As an action world model, MAPF-World integrates an action policy (fast system) and a world model (slow system) within a unified Transformer-based architecture. This framework employs a shared encoder and dual-branch decoders, comprising a fast head for rapid policy decisions and a slow head for future state prediction. We further introduce a spatio-agent positional encoding to facilitate decentralized coordination, 
capturing agent identities and relative motion intentions while distinguishing between static and dynamic spatio-temporal contexts.
Compared with reactive MAPF methods that directly map the current local observation to an action, 
MAPF-World additionally predicts the next local observation and neighboring agents' action intentions under supervised learning.
These predictions are autoregressively fed back into action selection, allowing agents to anticipate nearby interaction patterns and make more coordinated decisions under partial observability.

In summary, our main \textbf{contributions} are as follows:
\begin{itemize}
\item We propose MAPF-World, an autoregressive action--world model for MAPF that integrates action generation with short-horizon local future prediction for decentralized coordination under partial observation.
\item We present an automated pipeline for generating realistic grid-based MAPF maps from urban layouts, supporting configurable resolution, map size, and large-scale batch generation.
\item We conduct extensive experiments showing that MAPF-World improves decentralized coordination and zero-shot generalization across diverse map types and complex multi-agent scenarios.
% \item We conduct extensive experiments across diverse map families and interaction settings, showing that MAPF-World improves decentralized coordination and achieves robust zero-shot generalization compared with existing learned solvers.
\end{itemize}

\section{Related Work}

\subsection{Benchmarks} 
A problem instance is defined by three components: a map, a set of agents, and their start and goal positions. Existing benchmark maps~\cite{sturtevant2012benchmarks,stern2019mapf,skrynnik2025pogema} are generally grouped by topologies, including empty, random, mazes,  rooms\footnote{https://movingai.com/benchmarks/grids.html}\footnote{https://movingai.com/benchmarks/mapf.html},  warehouses\footnotemark[2] (12 maps), 
terrains\footnotemark[1] (20 maps),
video games\footnotemark[1] (454 maps),
and city streets\footnotemark[1]\footnotemark[2] (90 maps for 10 cities).
Recent works also investigate optimizing map layouts~\cite{zhang2023multi} and generating map patterns by neural cellular automata~\cite{zhang2023arbitrarily,qian2024quality}.
POGEMA~\cite{skrynnik2025pogema} includes a problem instance generator for Random, Mazes, and Warehouse maps.
However, real-world urban maps remain scarce, limiting the evaluation of solvers’ generalizability in practical settings. To address this, we introduce an automated pipeline for generating maps from real-world urban layouts.

\subsection{MAPF Solvers}
Classical MAPF solvers are typically search-based. Heuristic methods like CBS~\cite{sharon2015conflict}, PBS~\cite{liu2019task}, WHCA*~\cite{bnaya2014conflict} provide (sub)optimal solutions but struggle with scalability issues.
% with an increasing number of agents and map sizes. 
To address this, state-of-the-art (SOTA) anytime planners such as LaCAM*~\cite{okumura2023improving,okumura2023engineering} improve efficiency, 
offering a trade-off between speed and solution quality.
% For instance, CBS~\cite{sharon2015conflict} solves MAPF optimally by searching a conflict tree.
% PBS~\cite{liu2019task} is suboptimal by using a priority table to sequentially plan trajectories of agents.
% WHCA*~\cite{bnaya2014conflict} and RHCR~\cite{li2021lifelong} create suboptimal results by planning local paths within a time-windowed horizon.
% Despite their high solution quality, the above algorithms struggle to solve problems within a reasonable time as the number of agents and the map size increase.
% In this direction, PIBT~\cite{okumura2022priority} is a heuristic online planner suited for large-scale instances at the cost of solution quality.
% LaCAM*~\cite{okumura2023improving,okumura2023engineering} is a state-of-the-art anytime and near-optimal planner that addresses large-scale problems in a centralized and offline manner.
% Besides, for the lifelong multi-agent path-finding (LMAPF) problem, planners are challenged by the continuous new task allocation to agents reaching their goals. 
% RHCR~\cite{li2021lifelong} 
% For the lifelong multi-agent path-finding problem (LMAPF), agents need to be continuously assigned new tasks after completing a task. 
% The RHCR \cite{li2021lifelong} sequentially replans the paths of agents in a window, which the task allocator assigns a certain number of tasks to each agent to ensure that it always has tasks within the window, but it does not scale well. \cite{zhang2024guidance}. 
In contrast, learning-based methods offer scalable, efficient, and suboptimal solutions, primarily based on imitation learning (IL) and reinforcement learning (RL). 
These methods are generally categorized into hybrid and end-to-end approaches. Hybrid approaches replace partial components in heuristic algorithms with learnable modules. 
For example, 
% Follower~\cite{zhang2024guidance} first plans agents' paths using centralized heuristics, then executes them while avoiding collisions through a decentralized MARL policy.
When-to-Switch~\cite{skrynnik2023switch} combines memory-based replanning with a recurrent RL policy through a switching mechanism;
MCTS~\cite{skrynnik2024decentralized} uses Monte Carlo Tree Search and an RL policy for counterfactual planning;
LNS2+RL~\cite{wang2025lns2+} replaces the prioritized planning (PP) module in LNS2 with an RL strategy for improved sub-solution selection;
and LaGAT~\cite{jain2026graph} uses a graph neural network (GNN) to guide inter-agent priorities in LaCAM* and improves map-specific solution quality via fine-tuning.
The second category relies entirely on end-to-end learning. For example, 
% PRIMAL2~\cite{damani2021primal} combines imitation learning and MARL to train decentralized agent planners that avoid high traffic areas (e.g., corridors), thereby improving overall throughput.
SCRIMP~\cite{wang2023scrimp} uses MARL for communication and priority assignment to tackle challenges from agents' limited field of view, while MAPF-GPT~\cite{andreychuk2025mapf} introduces an 85M foundation model trained with IL on a 1B expert dataset, outperforming existing RL-based solvers. MAPF-GPT-DDG~\cite{andreychuk2025advancing} further improves MAPF-GPT via fine-tuning on difficult instances. 
% Despite these advances, most existing methods remain largely reactive and lack explicit future interaction modeling. 
% Our work addresses this gap by introducing an action-conditioned world model for more far-sighted planning.
% Despite these advances, prior learned decentralized MAPF methods either remain largely reactive or use explicit planning mechanisms, while MAPF-World integrates short-horizon prediction of future local observations and neighboring agents' intentions directly into decentralized action selection.
Despite these advances, prior learned decentralized MAPF methods either remain largely reactive or rely on explicit search/replanning mechanisms.
In contrast, MAPF-World integrates short-horizon prediction of future local observations and neighboring agents' action intentions directly into decentralized action selection.

\subsection{Action World Model and Fast--Slow Dual-System}
A world model~\cite{ha2018world,ha2018recurrent} is a generative model that predicts future states from current observations. 
With foundation models trained on large-scale data~\cite{bommasani2021opportunities}, world model learning for decision making has made significant strides in artificial general intelligence (AGI)~\cite{hafner2019dream,hafner2020mastering,hafner2025mastering,kanervisto2025world,ding2024understanding}.
Recently, the concept of action world model has been further introduced in WorldVLA~\cite{cen2025worldvla}, which extends traditional world models by integrating action generation into the world modeling and observation/action understanding into the action model. However, despite being an action world model, our proposed MAPF-World is built on Kahneman’s fast--slow dual-system~\cite{kahneman2011thinking}, akin to LeCun's two-mode architecture for the observation-action loop~\cite{lecun2022path}. Moreover, MAPF-World embeds the fast system in the slow system by sharing parameters, similar to the Fast-in-Slow (FiS) dual-system vision-language-action (VLA) model~\cite{chen2025fast}. A key distinction is that MAPF-World operates as an action world model, capable of autoregressively predicting the next local observation and multi-agent interactions required for multi-step planning.

\section{Method}

\begin{figure*}[htbp]
\centering
\includegraphics[width=0.95\textwidth]{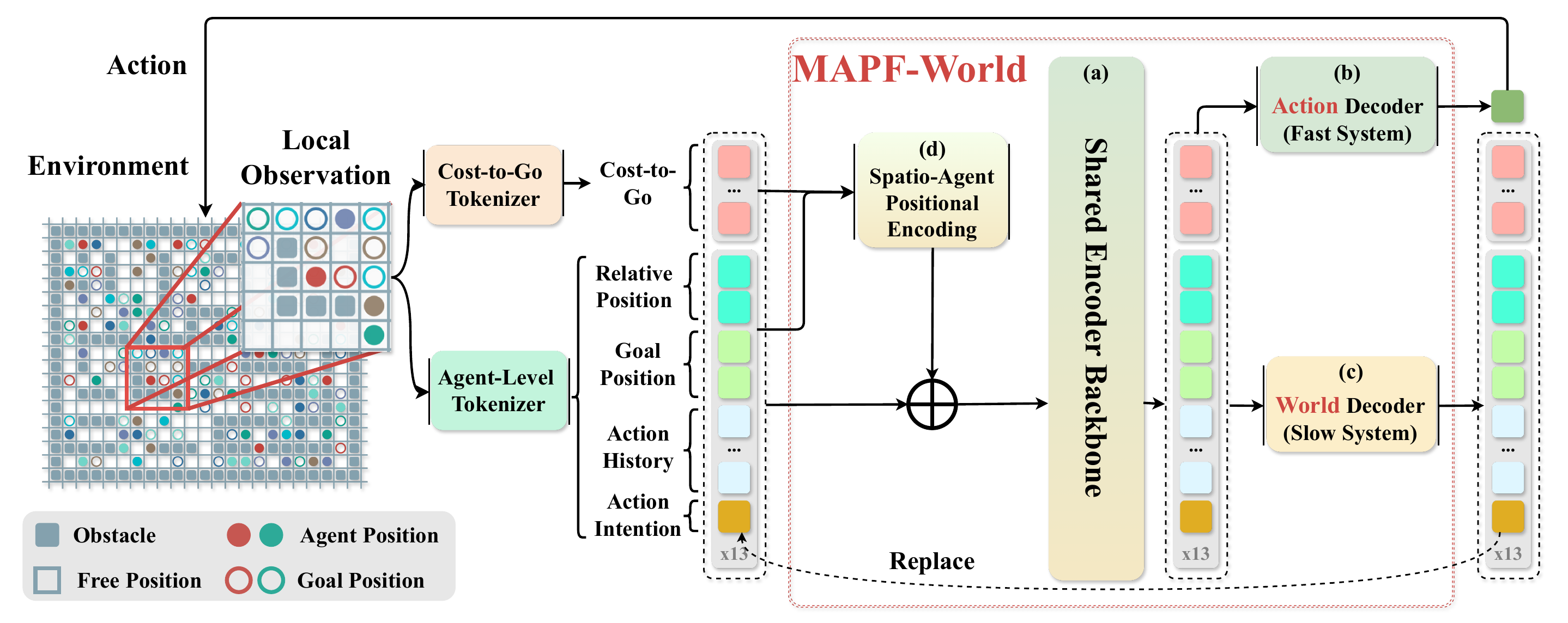}
\caption{Architecture of the proposed \textbf{MAPF-World}. 
Our model is designed as a \textbf{fast--slow dual-system} built on a hierarchical Transformer-based framework.
It features a shared \textbf{(a) backbone} for spatio-temporal representations, coupled with two dedicated components: an  \textbf{(b) action branch} (fast system) for rapid policy decisions and a \textbf{(c) world branch} (slow system) for local environmental transition prediction. Furthermore, we introduce an improved \textbf{(d) spatio-agent positional encoding} that distinguishes spatial guidance from the local map and semantic states at the agent level.}
\label{fig:actionworldmodel}
\end{figure*}

\begin{figure}[htbp]
\centering
\includegraphics[width=\linewidth]{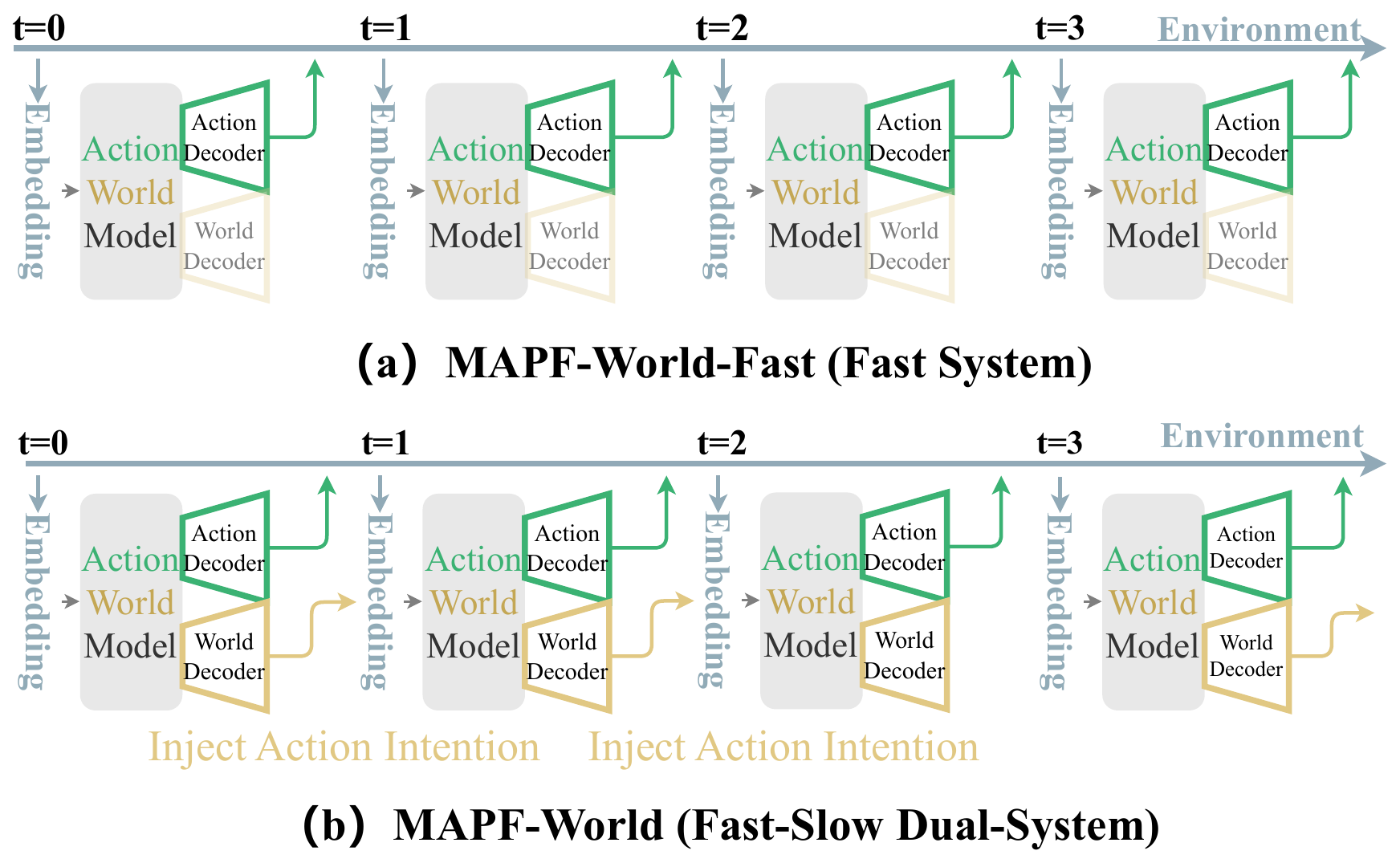}
\caption{Two operational modes of MAPF-World: (a) \textbf{MAPF-World-Fast}, a reactive system using only the action branch; (b) \textbf{MAPF-World}, a deliberative system integrating both action and world branches, where predicted and default greedy action intentions are interleaved to balance inter-agent coordination and goal-directed behavior.
}
\label{fig:mode_fast_slow}
\end{figure}
% \vspace{-10pt}

\begin{figure*}[htbp]
\centering
\includegraphics[width=0.98\textwidth]{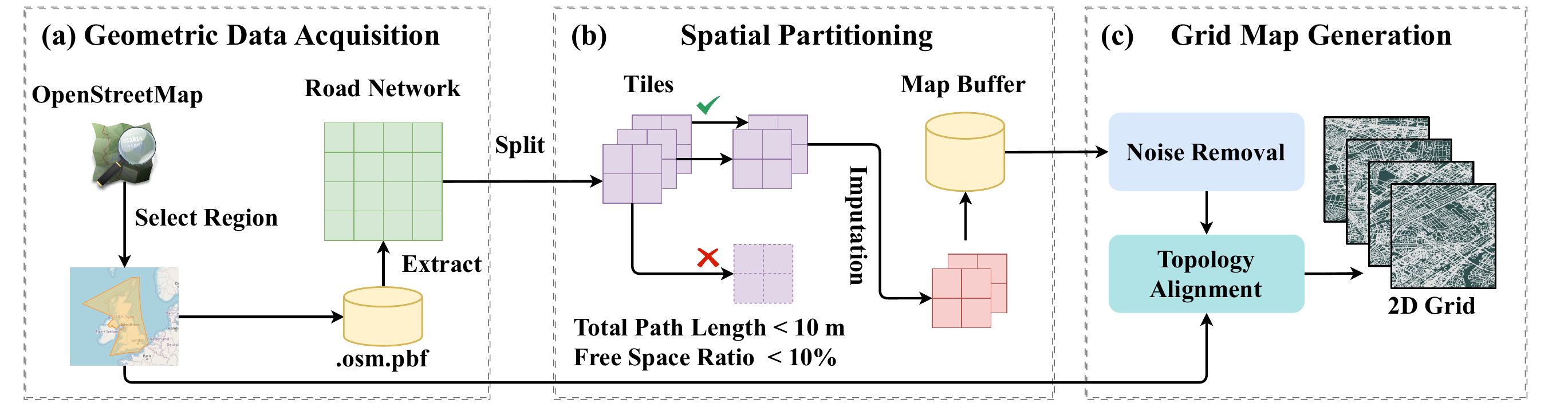}
\caption{Automated urban map generation pipeline.
(a) Geometric data acquisition: road networks are extracted from raw geographic data (\texttt{.osm.pbf}).
(b) Spatial partitioning: geographic space is partitioned into localized tiles; traversable regions and obstacles are identified with imputation of missing properties, and valid tiles are filtered based on configurable criteria.
(c) Grid map generation: tiles are rasterized into discrete occupancy grids, followed by denoising of isolated pixels and narrow gaps, and topological alignment to correct discretization artifacts.}
\label{fig:pipeline}
\end{figure*}

\subsection{Problem Formulation}
\label{sec:problem}
We consider a decentralized partially observable Markov decision process (Dec-POMDP)
with a set of agents $\mathcal{I} = \{1, \ldots, n\}$.
At time step $t$, each agent $i \in \mathcal{I}$ follows an action policy model
$\pi_\theta^i(a_t^i \mid o_t^i)$, which maps its local observation $o_t^i$ to an action $a_t^i$.
Each agent further maintains a world model
$f_\phi^i(\hat{o}_{t+1}^i \mid o_t^i)$ that captures the evolution of local observations, where $\hat{o}_{t+1}^i$ denotes the predicted observation at time step $t+1$, enabling explicit reasoning over local environmental dynamics beyond reactive policies.

We instantiate this formulation on the multi-agent path finding (MAPF) problem
over 2-D grid maps.
Given a set of start--goal position pairs $\{(\boldsymbol{p}_s^i, \boldsymbol{p}_g^i)\}_{i=1}^n$,
each agent selects an action
$a_t^i \in \{\text{Up}, \text{Right}, \text{Down}, \text{Left}, \text{Wait}\}$.
The resulting global joint action
$\boldsymbol{a}_t = \{a_t^1, \ldots, a_t^n\}$
induces the state transition
$\mathcal{P}(s_{t+1} \mid s_t, \boldsymbol{a}_t)$,
where $s_t$ encodes the joint agent positions.
A solution corresponds to a set of collision-free paths
$\{\tau^i\}_{i \in \mathcal{I}}$, with
$\tau^i = (\boldsymbol{p}_s^i, \boldsymbol{p}_1^i, \boldsymbol{p}_2^i, \ldots, \boldsymbol{p}_g^i)$.
We consider two types of collisions: vertex and edge collisions~\cite{stern2019mapf}.

\subsection{Unified Action--World Architecture}
\label{sec:architecture}

MAPF-World is an autoregressive action world model for MAPF that integrates action generation with short-horizon local future prediction, as shown in Fig.~\ref{fig:actionworldmodel}.
This formulation requires each agent to jointly model heterogeneous local information, including local map topology, neighboring-agent states, goal directions, action histories, and predicted action intentions.
These inputs have different semantics but interact within the same local decision context; for example, a neighboring agent's predicted intention is only meaningful when interpreted together with nearby obstacles, goal directions, and recent motion history.
We therefore represent each agent's $11\times11$ egocentric observation as a fixed-length structured token sequence and use self-attention to integrate cross-token dependencies among spatial, agent-level, and temporal coordination cues.

Built on this local-token representation, MAPF-World follows a fast--slow dual-system design with a shared Transformer backbone and two dedicated heads.
The {\bf fast system} (action head) generates the agent's next executed action, while the {\bf slow system} (world head) predicts the next local observation, including the inferred action intentions of the agent and its observable neighbors.
In particular, MAPF-World operates in two modes shown in Fig.~\ref{fig:mode_fast_slow}, allowing a trade-off between high-frequency reactivity and short-horizon predictive coordination.

\subsubsection{MAPF-World-Fast (Action Model)}
The fast system operates as a reactive policy $\pi_\theta^i(a_t^i \mid o_t^i)$ that leverages the shared Transformer backbone and action branch while bypassing the world branch during inference. It takes the current local observation $o_t^i$ as input to directly generate and execute the next action $a_t^i$, as illustrated in Fig.~\ref{fig:mode_fast_slow}~(a). This mode is designed for high-frequency responsiveness, prioritizing computational efficiency by focusing solely on immediate decision-making.

\subsubsection{MAPF-World (Action World Model)}
The fast--slow dual-system operates as an action--world model that integrates the action policy $\pi_\theta^i(a_t^i \mid o_t^i)$ and the world model $f_\phi^i(\hat{o}_{t+1}^i \mid o_t^i)$, as illustrated in Fig.~\ref{fig:mode_fast_slow}~(b).
The world branch predicts the next observation $\hat{o}_{t+1}^i$, while the action branch outputs the action $a_t^i$ for execution.
Specifically, $\hat{o}_{t+1}^i$ embeds inferred action intentions of agent $i$ and its observable neighbors.
For the default MAPF-World, we use a predictive horizon of $h=2$. Within each planning cycle, predicted action intentions are injected for the first $h$ consecutive steps, followed by greedy goal-oriented action intentions at step $h+1$. This interleaved autoregressive feedback mitigates accumulated predictive uncertainty while balancing goal-directed progress with inter-agent coordination.

\subsubsection{Spatio-Agent Positional Encoding (SAPE)}
Following prior work~\cite{andreychuk2025mapf}, the $11\times11$ local observation of the egocentric agent $i$ is structured into 256 tokens organized into three parts
using two tokenizers: a cost-to-go tokenizer and an agent-level tokenizer.
For the first part, the cost-to-go tokenizer computes an $11\times11$ local cost-to-go map consisting of 121 tokens, where each token encodes the distance reduction toward the goal when agent $i$ moves to the respective cell.
For the second part, the agent tokenizer generates 130 tokens encoding the 13 nearest agents in the local observation, sorted by their relative distances to agent $i$.
Each agent is represented by 10 tokens: a 2-D relative position with respect to agent $i$, a 2-D relative goal displacement (clamped to the local observation boundaries), a 5-D vector encoding the past five actions, and a scalar indicating the inferred \textbf{action intention}.
If fewer than 13 agents are present, the remaining tokens are padded with empty tokens.
The final five tokens are used for padding, yielding a total of 256 tokens.

We further introduce a spatio-agent positional encoding (\textbf{SAPE}) that jointly captures spatial and agent-level information.
In decentralized Transformer-based decision-making, a key challenge lies in injecting spatial awareness and agent-level semantics into the encoder.
Conventional positional encodings, which assign embeddings solely based on token indices, fail to capture agent identities and relative motion intentions.
SAPE addresses this limitation by aligning spatial guidance from the local cost-to-go map with semantic states encoded in agent-level tokens.

Specifically, the SAPE encoding is based on a basic operator
$\mathcal{E}_{\text{SA}}(x, y) : \mathbb{R}^2 \rightarrow \mathbb{R}^d$,
where $d$ denotes the model hidden dimension.
It first maps 2-D coordinate $(x, y)$ to $[r, \sin\theta, \cos\theta]$ to capture both distance and direction information, where 
$r = \sqrt{x^2 + y^2}$ and $\theta = \arctan2(y, x)$,
and then projects the resulting representation to a $d$-D vector through a linear layer.

The positional encoding for the cost-to-goal map tokens is denoted as
$\boldsymbol{E}_{\text{map}}^i =
[\mathcal{E}_{\text{SA}}(x_k, y_k)]_{k=1}^{121} \in \mathbb{R}^{121 \times d}$,
where $(x_k, y_k)$ denotes the local spatial position of the $k$-th cell.
For the agent-level tokens, the positional encoding for the neighboring agent $j$ observed by agent $i$ is denoted as
$\boldsymbol{E}^{i,j} \in \mathbb{R}^{10 \times d}$ and is defined as follows:
the first two tokens are assigned the relative position $\boldsymbol{p}^{i,j}$,
the next two tokens are assigned the goal position $\boldsymbol{p}_g^j$,
and the remaining six tokens are assigned the goal displacement
$\boldsymbol{p}_g^j - \boldsymbol{p}^{i,j}$.
This decomposition allows the model to explicitly distinguish static local spatial context from dynamic agent-level semantics, thereby facilitating more semantically informed attention allocation.

The final spatio-agent positional encoding for the 256 tokens is obtained by
$\text{Concat}(\cdot)$, combining the positional encodings of the cost-to-goal map,
the neighboring agents $j \in \mathcal{N}^i$ ($|\mathcal{N}^i| \leq 13$),
and five padding tokens:
\begin{equation}
\boldsymbol{E}_{\text{SA}}^i =
\text{Concat}\!\left(
\boldsymbol{E}_{\text{map}}^i,\,
\{\boldsymbol{E}^{i,j}\}_{j \in \mathcal{N}^i},\,
\boldsymbol{0}^{5 \times d}
\right)
\in \mathbb{R}^{256 \times d}.
\end{equation}

This positional encoding is added to the token embedding
$\boldsymbol{E}_{\text{Token}}^i$ before being fed into the Transformer encoder:
\begin{equation}
\boldsymbol{X}^i = \boldsymbol{E}_{\text{Token}}^i + \boldsymbol{E}_{\text{SA}}^i
\in \mathbb{R}^{256 \times d}.
\end{equation}

\subsection{Training Objective}
\label{sec:training}
MAPF-World is trained with two supervised objectives: fast-branch action prediction and slow-branch next-local-observation prediction.
The overall loss is
\begin{equation}
\mathcal{L}
=
\mathcal{L}_{\mathrm{act}}
+
\lambda \mathcal{L}_{\mathrm{pred}},
\qquad
\lambda = 0.5 .
\end{equation}
Here, $\mathcal{L}_{\mathrm{act}}$ is the cross-entropy loss on the final action token, while $\mathcal{L}_{\mathrm{pred}}$ is a weighted cross-entropy loss over the predicted tokens of the next local observation.
For $\mathcal{L}_{\mathrm{pred}}$, action-intention tokens are assigned weight $1.0$, padding tokens are ignored, and all remaining tokens are assigned weight $0.5$.
This auxiliary prediction loss encourages the shared representation of the backbone to capture local future interactions, while keeping action prediction as the primary training objective.

\subsection{Real-World Map Generator}
\label{sec:map_generator}
To bridge the gap between synthetic simulation and real-world deployment, we propose an automated pipeline for generating grid-based MAPF maps from real-world urban layouts.
The pipeline leverages OpenStreetMap (OSM)\footnote{https://www.openstreetmap.org/copyright} tags to identify traversable areas (e.g., footways, paths, residential roads, and pedestrian zones), while building footprints, natural barriers, and other non-navigable land-use areas are treated as obstacles.
It also supports batch processing over large geographic regions.
Using this pipeline, we generate 137 maps of size $256 \times 256$ from 11 European cities.

Specifically, as illustrated in Fig.~\ref{fig:pipeline}, the pipeline consists of three core stages:
(1) \textbf{Data Acquisition.} Road networks are extracted from \texttt{.osm.pbf} files as vector centerlines represented by geographic coordinates, with configurable physical resolution (e.g., 1\,m/grid) and map size (e.g., $100\times100$ grids).
Navigable road segments are selected based on map tags, and geographic boundaries are calibrated to isolate valid urban regions.
(2) \textbf{Spatial Partitioning.} The geographic space is partitioned into uniform spatial tiles with a configurable size. Within each tile, vector road segments are converted into polygonal representations of traversable space. When explicit width annotations are missing, road widths are estimated from available OSM tags (e.g., \texttt{highway}, \texttt{lanes}) using empirical defaults. Tiles are then filtered based on total path length (e.g., 10\,m) and free-space ratio (e.g., 10\%) to remove trivial or invalid areas.
(3) \textbf{Grid Map Generation.} The resulting vector polygons are rasterized into discrete grid maps. Isolated pixels and narrow gaps are removed via denoising, and the grid topology is aligned with the original geographic data (\texttt{.osm.pbf}) to correct discretization artifacts, yielding final grid-based maps suitable for MAPF.

We further compare EuropeanCity with existing benchmark families using a unified structural analysis pipeline (Table~\ref{tab:map_comparison}). EuropeanCity is not merely more challenging; it fills a benchmark gap by combining real urban layouts, configurable resolution, large-scale generation, narrow passages, and diverse intersections. Random and Mazes provide synthetic narrow-passage patterns, while MovingAI uses real city layouts but becomes relatively open after $64{\times}64$ rasterization. In contrast, our OSM-based pipeline can generate maps from arbitrary geographic regions while preserving fine-grained road-network topology.

Compared with MovingAI, EuropeanCity has a $4.9\times$ higher narrow-passage ratio ($0.85$ vs.\ $0.17$), $2.6\times$ more connected components ($10.1$ vs.\ $3.9$), and nearly $5\times$ narrower corridors ($2.6$ vs.\ $12.9$ cells). Its local topology is also more heterogeneous, containing corridors, T-junctions, and 4-way crossings, rather than being dominated by open degree-4 cells. These properties make EuropeanCity a complementary benchmark for evaluating MAPF solvers under realistic and structurally diverse urban morphologies.

\begin{table}[t]
\centering
\caption{Structural statistics of MAPF benchmark map families. All metrics are computed using the same pipeline.}
\label{tab:map_comparison}
\renewcommand{\arraystretch}{1.10}
\setlength{\tabcolsep}{3pt}
\resizebox{\columnwidth}{!}{%
\begin{tabular}{lccccc} 
\toprule 
\textbf{Metric} & \textbf{Random} & \textbf{Mazes} & \textbf{Warehouse} & \textbf{MovingAI} & \textbf{EuropeanCity} \\ 
& $(n{=}128)$ & $(n{=}128)$ & $(n{=}1)$ & $(n{=}128)$ & \textbf{(Ours, $n{=}137$)} \\ 
\midrule Size & $17^2$--$21^2$ & $17^2$--$21^2$ & $33{\times}46$ & $64^2$ & $\mathbf{256^2}$ \\
Free Ratio & $0.80{\pm}0.06$ & $0.70{\pm}0.10$ & $0.84$ & $0.73{\pm}0.09$ & $0.46{\pm}0.11$ \\ 
CC & $2.7{\pm}1.9$ & $1.0{\pm}0.0$ & $1.0$ & $3.9{\pm}3.6$ & $\mathbf{10.1{\pm}7.3}$ \\ 
NPR & $0.78{\pm}0.12$ & $0.79{\pm}0.27$ & $0.45$ & $0.17{\pm}0.07$ & $\mathbf{0.85{\pm}0.09}$ \\ 
MCW & $2.76{\pm}0.41$ & $3.59{\pm}4.11$ & $4.82$ & $12.9{\pm}8.3$ & $\mathbf{2.61{\pm}0.47}$ \\ 
\bottomrule 
\end{tabular}
}
\par\vspace{1mm}
\begin{minipage}{\columnwidth}
\footnotesize
\raggedright
\textit{Note:} Free Ratio denotes the fraction of traversable cells; 
CC denotes the number of 4-connected free-cell components; 
NPR denotes the fraction of free cells within ${\leq}1.5$ cells of the nearest obstacle; 
MCW denotes the mean corridor width in cells.
\end{minipage}
\end{table}

\begin{figure*}[htbp]
    \centering
    \includegraphics[width=\textwidth]{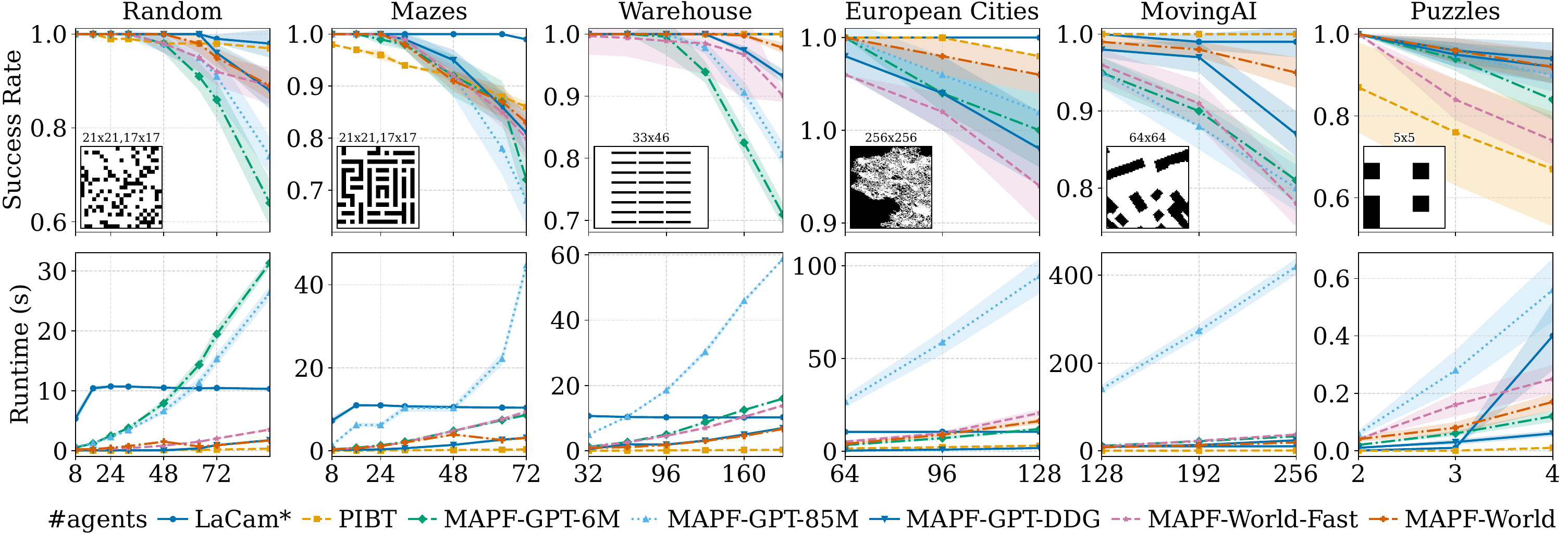}
    \vspace{-1.5em}
    \caption{Success rate and runtime evaluation across benchmark maps. Random and Mazes are in-distribution, while the remaining maps are out-of-distribution (OOD). Among learning-based solvers, MAPF-World shows the smallest performance degradation and near-linear runtime scaling as agent density increases, indicating stronger zero-shot generalization.}
    \label{fig:compare_baseline}
\end{figure*}

\begin{figure}[htbp]
    \centering
    \includegraphics[width=0.5\textwidth]{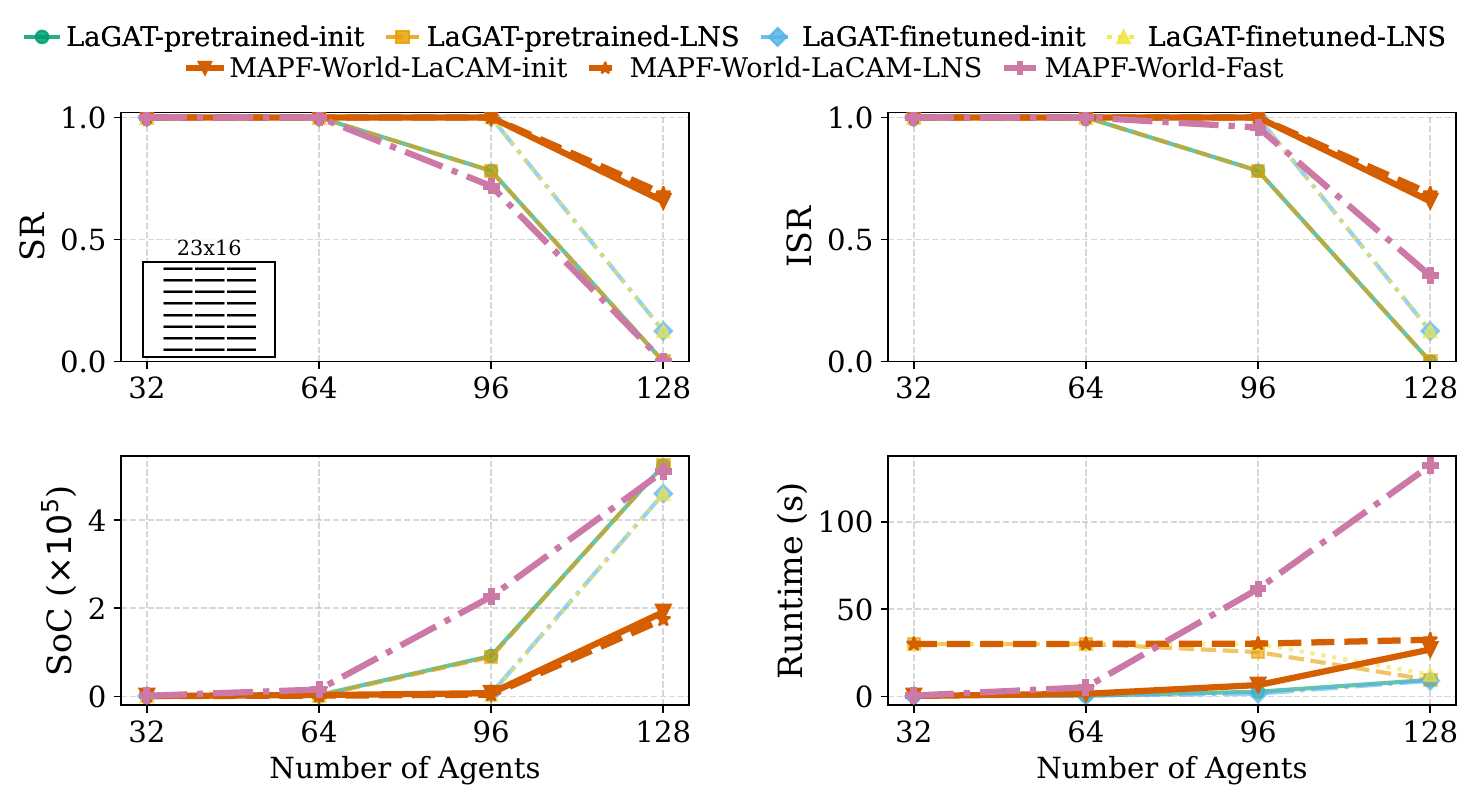}
    \vspace{-1.5em}
    \caption{Comparison between GNN-guided LaGAT variants and MAPF-World variants on the dense warehouse setting.}
    \label{fig:compare_LaGAT}
\end{figure}

\begin{figure}[htbp]
\includegraphics[width=\linewidth]{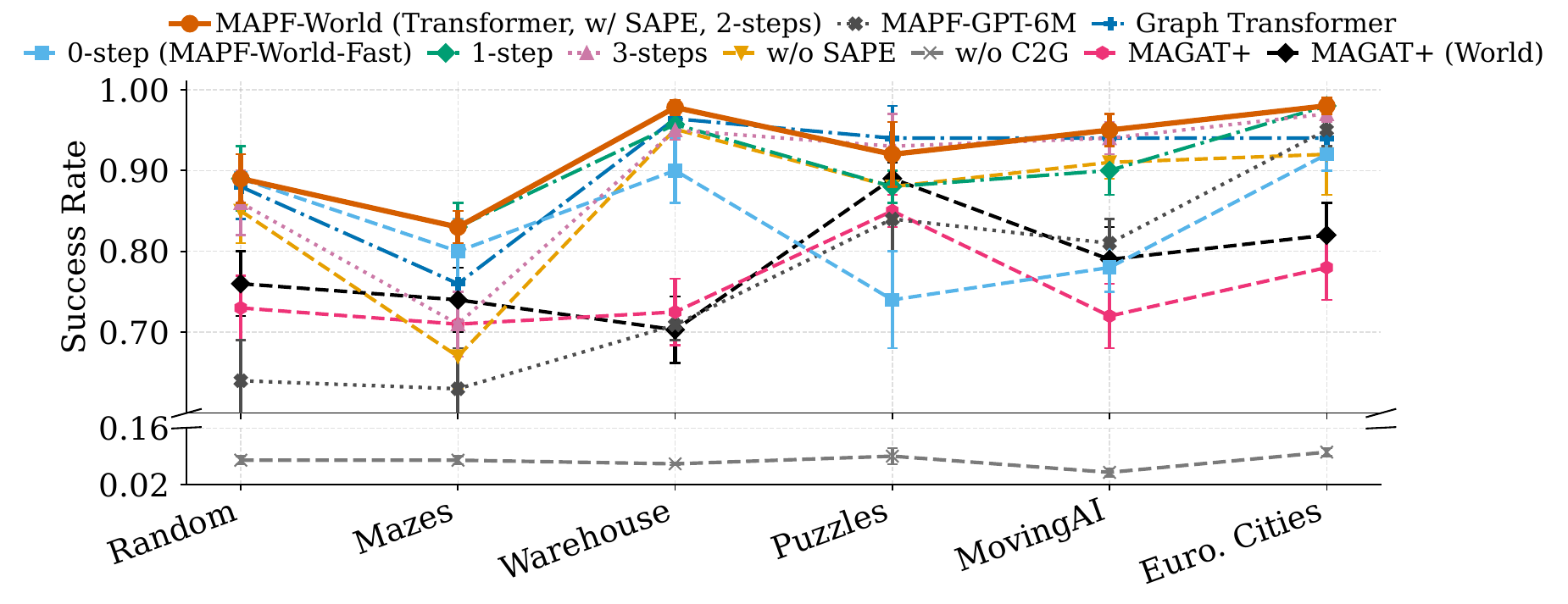}
\caption{Ablation results of MAPF-World, including prediction horizon, SAPE, C2G tokens, and backbone architecture.}
\label{fig:ablation_all}
\end{figure}

\begin{figure}[htbp]
    \centering
    \includegraphics[width=\linewidth]{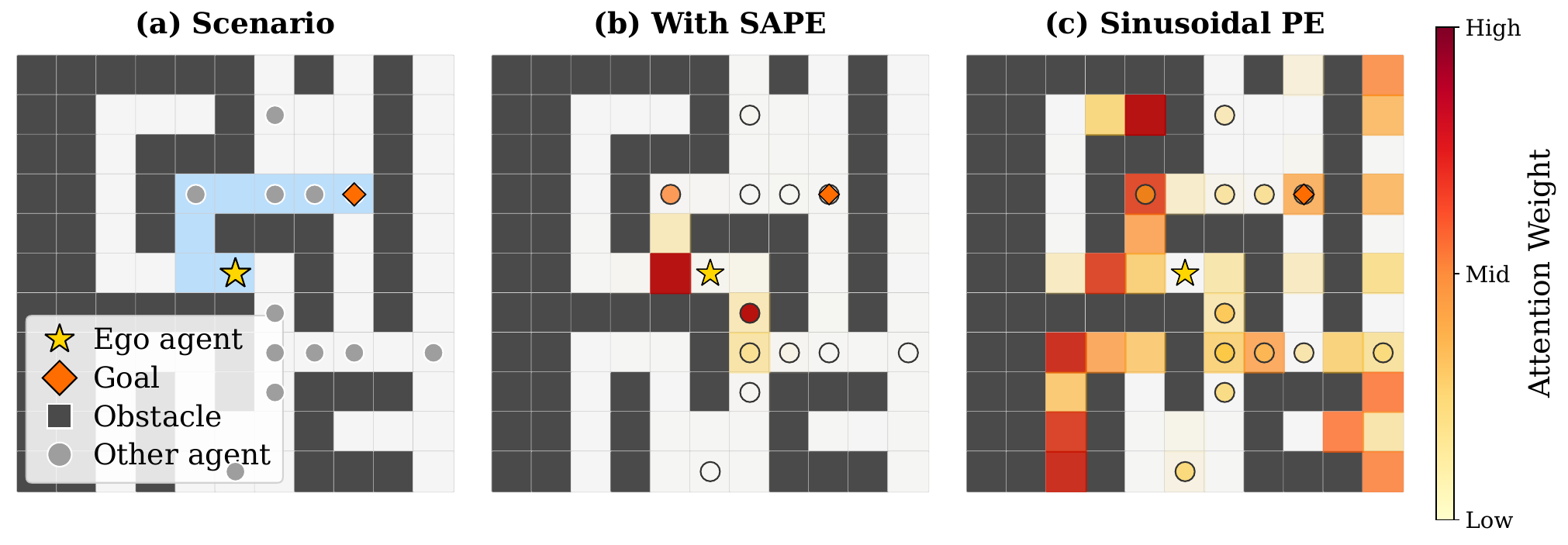}
    \caption{Attention comparison between SAPE and sinusoidal PE within the local observation window. SAPE highlights nearby agents and coordination-critical regions.}
    \label{fig:visualize_attention}
\end{figure}

\section{Experiments}
We conduct extensive experiments to address the following research questions:
(1) How does MAPF-World compare with existing MAPF solvers?
(2) Does the action--world model formulation improve decentralized coordination?
(3) Which architectural and representation components contribute most to performance, and under what scenarios does the predictive branch still struggle?

\subsection{Datasets}
MAPF-World's fast head is trained on the same 1B one-step observation--action pairs $(o_t, a_t)$ as MAPF-GPT~\cite{andreychuk2025mapf}. 
For the slow head, we generate 80M transition tuples $(o_t, o_{t+1}, a_t)$ (40M each from Random and Mazes maps) using the same MAPF-GPT data generation pipeline. 
Specifically, we use LaCAM*~\cite{okumura2023engineering} to solve 300K problem instances, corresponding to 8\% of the 3.75M instances used in MAPF-GPT, with a 10~s time limit. 
Unlike MAPF-GPT, we retain ``wait-at-target'' actions, since they are integral to the planning sequences and thus long-horizon planning.

\subsection{Benchmarks and Metrics}

For evaluation, we compare MAPF solvers on six map families: Random, Mazes, Warehouse, Puzzles, MovingAI, and EuropeanCity maps, as summarized in Table~\ref{tbl_datasets_test}. The first five map families are provided by POGEMA~\cite{skrynnik2025pogema}. The EuropeanCity maps are generated by our proposed map generator, with examples shown in Fig.~\ref{fig:pipeline}.
We report standard MAPF metrics: cooperative success rate (SR), individual success rate (ISR), sum of costs (SoC), and runtime, corresponding to solved-instance ratio, successful-agent ratio, total agent travel time steps, and wall-clock computation time, respectively.

\begin{table}[htbp!]
\centering
\footnotesize
\setlength{\tabcolsep}{3pt}
\caption{Details of benchmarks used in evaluation.}
\label{tbl_datasets_test}
\begin{tabular}{c|c|c|c|c|c}
\toprule
Dataset
& Agents
& Maps
& Map Size
& Seeds
& Max Steps
\\ \hline\hline
Random
% & \tabincell{c}{8, 16, 24, 32,\\ 48, 64, 72, 96}
& 8, 16, $\ldots$, 96
& 128
& 17$\times$17, 21$\times$21
& 1
& 128
\\ 
% \hline
Mazes
% & \tabincell{c}{8, 16, 24, 32,\\ 48, 64, 72}
& 8, 16, $\ldots$, 72
& 128
& 17$\times$17, 21$\times$21
& 1
& 128
\\ 
% \hline
Warehouse
% & \tabincell{c}{32, 64, 96, 128,\\ 160, 192}
& 32, 64, $\ldots$, 192
& 1
& 33$\times$46
& 128
& 128
\\ 
% \hline
MovingAI
& 128, 192, 256
& 128 
& 64$\times$64
& 1
& 256
\\ 
% \hline
Puzzles
& 2, 3, 4
& 16
& 5$\times$5
& 10
& 128
\\ 
% \hline
European
% (downsample)
& 64, 96, 128
& 137
& 256$\times$256
& 1
& 128
% \\ 
% \hline
% European
% & 128, 192, 256
% & 137
% & 256$\times$256
% & 1
% & 512
\\ \bottomrule
\end{tabular}
\end{table}

\subsection{Baselines and Model Variants}
\begin{itemize}
\item LaCAM*~\cite{okumura2023improving,okumura2023engineering}: a scalable SOTA centralized anytime heuristic method with high efficiency.
\item PIBT~\cite{okumura2022priority}: a scalable centralized online heuristic method for efficient collision avoidance.
\item MAPF-GPT-85M~\cite{andreychuk2025mapf}: a large foundation model for decentralized MAPF with 85M parameters (45.9 GFLOPs) trained on a 1B expert dataset via imitation learning.
\item MAPF-GPT-6M~\cite{andreychuk2025mapf}: a scaled-down variant of MAPF-GPT with 6M parameters (3.8 GFLOPs), trained on a reduced dataset of 150M samples.
\item MAPF-GPT-DDG~\cite{andreychuk2025advancing}: a SOTA MAPF solver fine-tuned on MAPF-GPT-2M model (1 GFLOPs), reported to surpass all contemporary learnable solvers.
\item LaGAT~\cite{jain2026graph}: a SOTA hybrid method for decentralized MAPF combining GNN-based learning with LaCAM*-based search.
% \item MAPF-World-GT (ours): a backbone ablation replacing the Transformer with a Graph Transformer~\cite{yun2019graph} under the same action-world framework.
\item MAPF-World (ours): an action world model based on a fast--slow dual-system, with 2.7M parameters (1.8 GFLOPs) trained on our collected 80M-sample dataset.
\item MAPF-World-Fast (ours): the fast system of MAPF-World, with 1.5M parameters (1 GFLOPs) trained on the same 1B dataset as MPAF-GPT-85M.
\end{itemize}

\subsection{Implementations}
Generating the 80M expert dataset takes about two days on a 16-core CPU workstation. 
Training MAPF-World on Random and Mazes takes 48 hours on one NVIDIA V100 GPU with 64 CPU cores; the remaining map families are used only for zero-shot inference. 
Following MAPF-GPT~\cite{andreychuk2025mapf}, we set the hidden dimension to $d=160$.

\subsection{Main Results}

\textit{Ours vs. SOTA Baselines.}
Fig.~\ref{fig:compare_baseline} compares MAPF-World with classical, heuristic, and learned MAPF solvers across six benchmark map families. Overall, MAPF-World achieves the strongest performance among learned solvers on most out-of-distribution scenarios, especially Warehouse, MovingAI, Puzzles, and EuropeanCity maps, while maintaining competitive runtime. Compared with MAPF-GPT-85M and MAPF-GPT-DDG, MAPF-World shows smaller performance degradation as agent density increases, indicating stronger zero-shot cross-map generalization. The advantage is more pronounced in complex or congested scenarios, where short-horizon prediction of future local observations and neighboring agents' action intentions helps improve coordination.

We further compare with LaGAT, a representative GNN-guided hybrid MAPF solver, on its released dense warehouse setting on an NVIDIA RTX 4090D GPU in Fig.~\ref{fig:compare_LaGAT}. Since LaGAT is map-specific and can fine-tune its learning component MAGAT+ for individual maps, we report this comparison separately from our full cross-map benchmark. Without map-specific fine-tuning, zero-shot MAPF-World-Fast improves over pretrained LaGAT and approaches map-specific fine-tuned LaGAT when used as the learned guidance module in the same LaCAM/LNS~\cite{li2021anytime} framework, especially under high agent densities. LaGAT achieves lower runtime due to its lightweight C++ hybrid implementation, while MAPF-World shows stronger zero-shot generalization in dense warehouse scenarios.

\textit{Fast-only vs. Fast--Slow System.}
A direct comparison between MAPF-World-Fast and MAPF-World reveals the effect of the slow predictive branch. As shown in Fig.~\ref{fig:compare_baseline}, MAPF-World consistently outperforms MAPF-World-Fast across nearly all test cases, with the largest gains on more complex maps such as MovingAI, Puzzles, and European City maps. Meanwhile, MAPF-World-Fast remains competitive in low-density settings, such as Random and Warehouse maps, or in locally navigable layouts such as Mazes. These results show that the slow branch improves predictive situational awareness and coordination, while the fast-only system remains effective in less congested scenarios.

Although the per-step neural forward cost of MAPF-World is 1.799 GFLOPs, about $1.8\times$ that of MAPF-World-Fast, MAPF-World often achieves lower episode runtime due to its improved decentralized coordination.

\subsection{Ablation Study}

We conduct ablations on predictive horizon, SAPE, backbone architecture, and token components to isolate the key factors of MAPF-World, as shown in Fig.~\ref{fig:ablation_all}.

\textit{Predictive horizon.}
We vary the number of predictive steps before one greedy goal-oriented step. The 0-step variant is MAPF-World-Fast, while the 2-step setting refers to the default MAPF-World and achieves the best overall SR. The weaker 3-step result indicates that longer predictive rollout may accumulate errors beyond useful future modeling.

\textit{Spatio-agent positional encoding.}
Removing SAPE causes a $3\%$--$19\%$ SR drop ($6.6\%$ on average), confirming the benefit of distinguishing static spatial context from dynamic agent-level information. 
Fig.~\ref{fig:visualize_attention} further shows that SAPE leads to stronger attention to nearby agents and coordination-critical regions than sinusoidal positional encoding (PE).

\textit{Backbone architecture.}
We replace the dense local Transformer with a Graph Transformer~\cite{dwivedi2020generalization} and a message-passing GNN method MAGAT+\cite{jain2026graph} under the same action-world objective. 
The Transformer backbone generally performs better, while action-world variants outperform their action-only counterparts and MAPF-GPT-6M on most map families.
This suggests the beneficial applicability of the action-world formulation to different backbone structures, while backbone architecture materially affects the coordination performance.

\textit{Token components.}
Cost-to-go (C2G) is the most critical token component: removing goal or action-history tokens preserves an average SR of $78\%$, while removing greedy-action tokens gives $54\%$. Removing C2G tokens collapses the average SR to $7\%$, far below MAPF-GPT-6M's $23\%$~\cite{andreychuk2025mapf}, showing that cost-to-go guidance is essential for grounding short-horizon prediction in goal-directed local topology.

\subsection{Failure Analysis}

\begin{figure}[htbp]
\centering
\includegraphics[width=\linewidth]{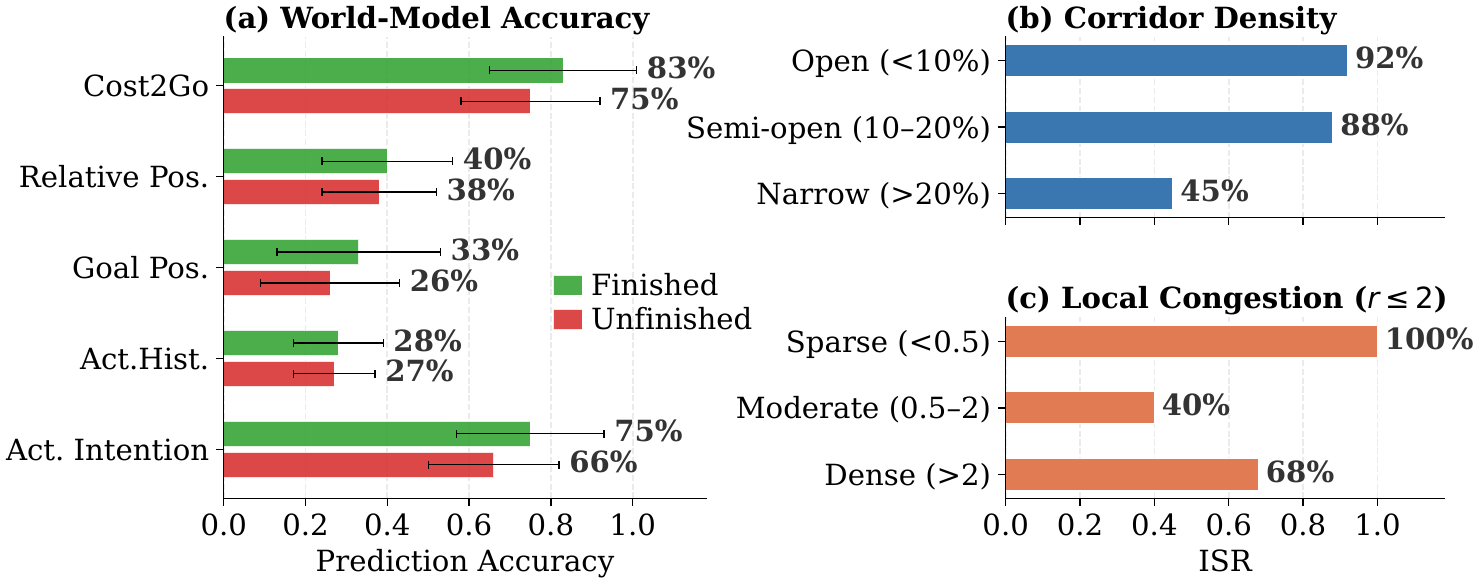}
\caption{Failure analysis of MAPF-World-Slow. 
(a) Next-step prediction accuracy by token segment for finished and unfinished agents. 
(b) ISR under different corridor densities. 
(c) ISR under different local congestion levels.}
\label{fig:failure_analysis}
\end{figure}

Failed episodes typically end by timeout after local deadlock-like, livelock-like, or slow-progress behaviors. 
Fig.~\ref{fig:failure_analysis}(a) shows that unfinished agents have lower next-step prediction accuracy, especially on cost-to-go tokens (75\% vs. 83\%) and action-intention tokens (66\% vs. 75\%), linking prediction errors to task failure.
Fig.~\ref{fig:failure_analysis}(b) shows that ISR drops from 92\% in open maps to 45\% in narrow-corridor maps, where multi-step yielding and priority negotiation may exceed the short prediction horizon.
Fig.~\ref{fig:failure_analysis}(c) shows that ISR decreases in non-sparse neighborhoods but not monotonically, suggesting that failures arise from the interaction of congestion, topology, short-horizon prediction, and decentralized local observability.

\section{Conclusion}
In this work, we propose MAPF-World, an autoregressive action--world model for decentralized MAPF under partial observation. By combining fast action generation with short-horizon local future prediction, MAPF-World enables agents to anticipate nearby interactions beyond purely reactive policies. Experiments show improved success rates and robust zero-shot generalization across maps among learned solvers, especially in crowded and structurally complex scenarios, with moderate computational overhead. Ablations validate the contributions of the predictive horizon, SAPE, backbone design, and token components, while failure analysis highlights remaining challenges in narrow passages, local congestion, and limited short-horizon prediction. We also introduce an automated map-generation pipeline from real-world urban layouts, providing a structurally diverse benchmark for evaluating robust and generalizable MAPF solvers.

% \section{Acknowledgments}

% \bibliographystyle{apalike}
% \bibliographystyle{plainnat}
\bibliographystyle{IEEEtran}
\bibliography{references}

% \clearpage
% \onecolumn
% \appendix

\end{document}